\pdfoutput=1

\documentclass[11pt]{article}

\usepackage[]{ACL2023}

\usepackage{times}
\usepackage{latexsym}
\usepackage{color}
\usepackage{framed}
\definecolor{shadecolor}{rgb}{0.92,0.92,0.92}
\usepackage{amsmath}

\usepackage{array}
\usepackage{utfsym}
\usepackage{subfigure}
\usepackage{balance}
\usepackage[misc]{ifsym}
\usepackage{url}
\usepackage{graphicx}
\usepackage{enumitem}
\usepackage{makecell}
\usepackage{booktabs}
\usepackage{multirow}
\usepackage{amssymb}
\usepackage{color,xcolor}
\usepackage{cases}
\usepackage{colortbl} 
\usepackage{diagbox}

\definecolor{ggreen}{HTML}{d4e7cf}
\usepackage[T1]{fontenc}

\usepackage[utf8]{inputenc}

\usepackage{microtype}

\title{Is ChatGPT a Good NLG Evaluator? A Preliminary Study}

\author{Jiaan Wang\textsuperscript{1}\thanks{\ \ Equal Contribution. Work was done when Wang and Liang was interning at Pattern Recognition Center, WeChat AI, Tencent Inc, China.}\ \thanks{ \ \ Corresponding author.}, \ Yunlong Liang\textsuperscript{2}\footnotemark[1], \ Fandong Meng\textsuperscript{3} \\
\bf{Zengkui Sun\textsuperscript{2}}, \ Haoxiang Shi\textsuperscript{4}, \ Zhixu Li\textsuperscript{5}, \ Jinan Xu\textsuperscript{2}, \ Jianfeng Qu\textsuperscript{1} and Jie Zhou\textsuperscript{3} \\
\textsuperscript{1}Soochow University, Suzhou, China \quad \textsuperscript{2}Beijing Jiaotong University, Beijing, China  \\
\textsuperscript{3}Pattern Recognition Center, WeChat AI, Tencent Inc, China \quad \textsuperscript{4}Waseda University, Tokyo, Japan \\
\textsuperscript{5}Fudan Unversity, Shanghai, China \\
 \\ 
\texttt{jawang.nlp@gmail.com} \quad \texttt{\{yunlongliang, zengksun\}@bjtu.edu.cn} \\ 
\texttt{fandongmeng@tencent.com} \quad \texttt{hollis.shi@toki.waseda.jp}
}

\begin{document}
\maketitle
\begin{abstract}
Recently, the emergence of ChatGPT has attracted wide attention from the computational linguistics community. Many prior studies have shown that ChatGPT achieves remarkable performance on various NLP tasks in terms of automatic evaluation metrics. However, the ability of ChatGPT to serve as an evaluation metric is still underexplored.
Considering assessing the quality of natural language generation (NLG) models is an arduous task and NLG metrics notoriously show their poor correlation with human judgments, we wonder \emph{whether ChatGPT is a good NLG evaluation metric.}

In this report, we provide a preliminary meta-evaluation on ChatGPT to show its reliability as an NLG metric.
In detail, we regard ChatGPT as a human evaluator and give task-specific (\emph{e.g.}, summarization) and aspect-specific (\emph{e.g.}, relevance) instruction to prompt ChatGPT to evaluate the generated results of NLG models.
We conduct experiments on five NLG meta-evaluation datasets (including summarization, story generation and data-to-text tasks).
Experimental results show that compared with previous automatic metrics, ChatGPT achieves state-of-the-art or competitive correlation with human judgments in most cases.
In addition, we find that the effectiveness of the ChatGPT evaluator might be influenced by the creation method of the meta-evaluation datasets.
For the meta-evaluation datasets which are created greatly depending on the reference and thus are biased, the ChatGPT evaluator might lose its effectiveness.
We hope our preliminary study could prompt the emergence of a general-purposed reliable NLG metric.\footnote{We have released the used data at \url{https://github.com/krystalan/chatgpt_as_nlg_evaluator}.}

\end{abstract}

\begin{figure}[t]
\centerline{\includegraphics[width=0.48\textwidth]{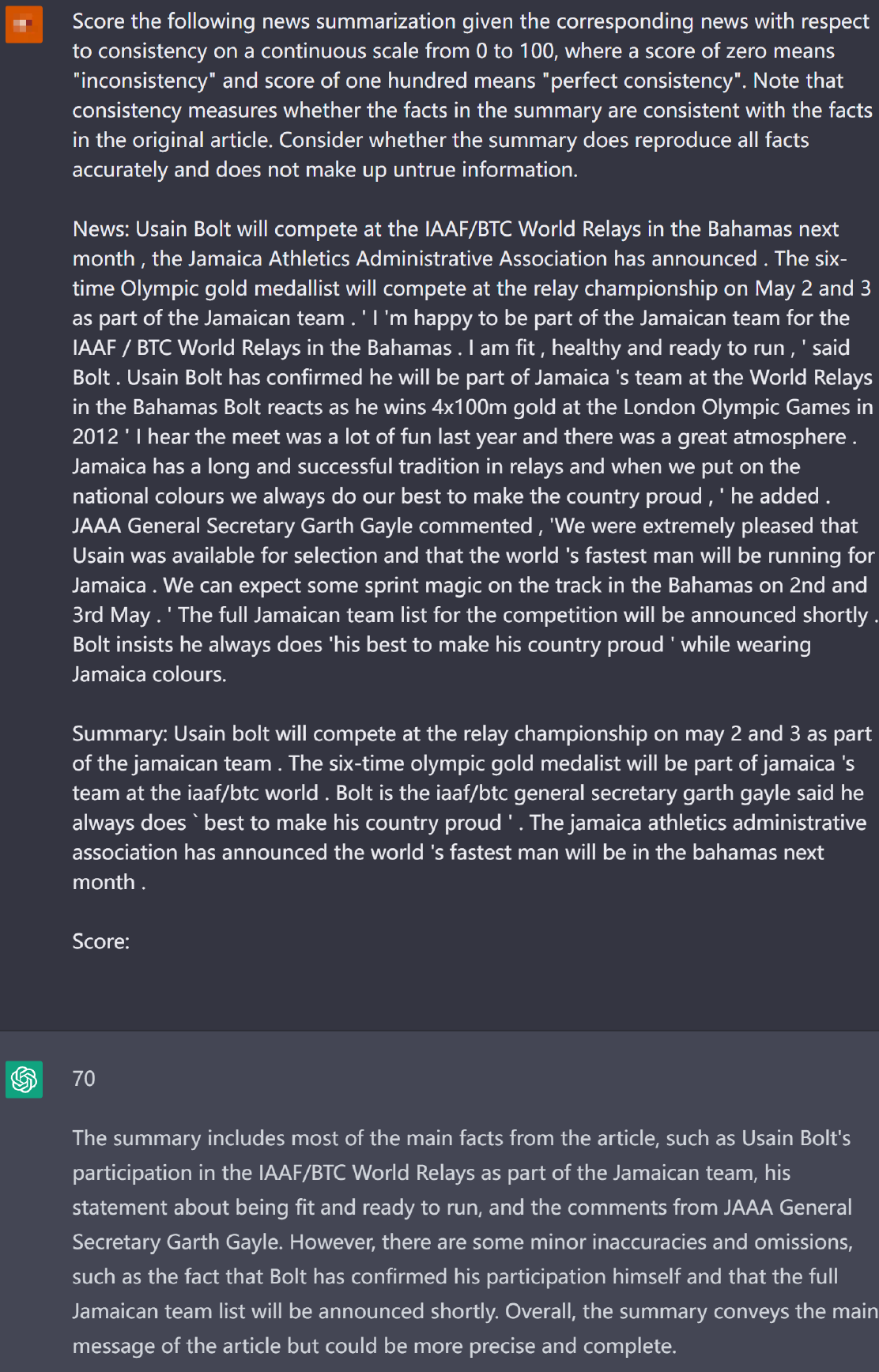}}
\caption{Prompting ChatGPT as an evaluator to score the generated results of NLG models (taking news summarization as an example).}
\label{fig:example}
\vspace{-0.5cm}
\end{figure}

\section{Introduction}
Pre-trained large language models (LLMs; \emph{e.g.}, GPT-3.5, ChatGPT and GPT-4), which are performed through chatting (or asking) with it, have obtained promising results on various natural language understanding (NLU) and natural language generation (NLG) downstream tasks~\cite{ouyang2022training,https://doi.org/10.48550/arxiv.2302.10724,qin2023chatgpt,https://doi.org/10.48550/arxiv.2302.07736,Yang2023ExploringTL,https://doi.org/10.48550/arxiv.2303.01248,bang2023multitask,https://doi.org/10.48550/arxiv.2302.13793}.
For example, \citet{https://doi.org/10.48550/arxiv.2302.10198} show that ChatGPT can attain the comparable understanding ability to some fine-tuned BERT-style models on NLU tasks while failing to surpass current task-specific NLU models.
\citet{https://doi.org/10.48550/arxiv.2302.10205} prove that ChatGPT can achieve good performance and even surpasses some full-shot models on several datasets through a multi-turn question-answering manner.
For NLG tasks, \citet{jiao2023chatgpt} claim that ChatGPT performs competitively with commercial translation products (\emph{e.g.}, Google Translator) on high-resource European languages.
\citet{wang2023cross} demonstrate that ChatGPT can balance well between informativeness and conciseness, and generate great cross-lingual summaries. Although impressive performance on these tasks in terms of automatic evaluation metrics has been shown, it is still not clear \emph{whether ChatGPT can evaluate the quality of textual generations as a human does.}

Recently, using pre-trained language models as NLG evaluation metric, \emph{e.g.}, MoverScore~\cite{zhao-etal-2019-moverscore}, BERTScore~\cite{Zhang2020BERTScoreET}, COMET~\cite{rei-etal-2020-comet}, BLEURT~\cite{sellam-etal-2020-bleurt}, BARTScore~\cite{https://doi.org/10.48550/arxiv.2106.11520} and MAUVE~\cite{https://doi.org/10.48550/arxiv.2212.14578}, receives increasing attention since it offers a decent human-related judgment from a deep semantic perspective. Given the powerful ability of ChatGPT as an intelligent conversational LLM, researchers also attempt to investigate whether it can evaluate the translation quality as a human evaluator~\cite{kocmi2023large}. However, the automated assessment of the general generation quality of NLG models still remains underexplored. 

In this report, we aim to answer the following research question: \emph{Is ChatGPT a good NLG evaluator?} To this end, we regard ChatGPT as a human evaluator and give task-specific (\emph{e.g.}, summarization) and aspect-specific (\emph{e.g.}, relevance) instruction to prompt ChatGPT to evaluate the generation of NLG models.
As the example shows in Figure~\ref{fig:example}, we also attempt different scoring criteria and whether to provide golden references in the prompts to systematically test the reliability of the ChatGPT evaluator. 
We conduct experiments on five widely-used NLG meta-evaluation datasets (including summarization, story generation and data-to-text tasks).
Experimental results show that ChatGPT exhibits a high correlation with human judgment in most cases especially for the story generation task, indicating its potential as an NLG metric.
In addition, we find that the ChatGPT evaluator is sensitive to the prompts, and for different tasks or aspects, the prompts should be carefully designed.
Moreover, the creation method of the meta-evaluation datasets has a significant influence on the effectiveness of different evaluation metrics.
If a meta-evaluation dataset is created greatly depending on the reference, the similarity between model generation and references serves as a strong signal to reflect human judgments, where simple similarity-based metrics (\emph{e.g.}, ROUGE) can achieve very strong performance. Therefore, the ChatGPT evaluator might lose its effectiveness in such situations.

Our main contributions are concluded as follows:
\begin{itemize}[leftmargin=*,topsep=0pt]
\setlength{\itemsep}{0pt}
\setlength{\parsep}{0pt}
\setlength{\parskip}{0pt}
\item To our knowledge, we are the first to utilize ChatGPT as a general NLG evaluation metric to study its correlations with human judgments.
\item We use task-specific and aspect-specific prompts to guide ChatGPT to perform as a reference-free or reference-based NLG metric, and evaluate its effectiveness on five widely-used meta-evaluation datasets covering three NLG tasks.
\item We find that the ChatGPT evaluator has a high correlation with humans in most cases, especially for creative NLG tasks (\emph{e.g.}, story generation) where multiple generations can satisfy humans.
\item We find that the ChatGPT evaluator is sensitive to the prompts. For different tasks and aspects, the prompt should be carefully designed.
\item We find that the involved biases of the NLG meta-evaluation datasets also influence the effectiveness of NLG metrics, and might lead to the limited effectiveness of the ChatGPT evaluator.
\end{itemize}

\section{Related Work}
\subsection{NLG Metrics}
A good automatic NLG metric can effectively indicate the quality of the textual generations and thus can save lots of human labor from conducting human evaluation. Therefore, it is vital to design automatic evaluation metrics for NLG tasks, \emph{e.g.}, text summarization, story generation, data-to-text generation, machine translation, and many others. Generally, the score that points out how well the systems perform on each task is computed by comparing the system texts with one or more reference texts for semantic matching. In the literature, the metrics can be roughly categorized into four types:

\paragraph{$n$-gram-based Metrics.}
Essentially, the $n$-gram-based metrics aim to measure the lexical overlap between a generated text and a reference text. The standard $n$-gram overlap-based metrics generally include ROUGE~\cite{lin-2004-rouge}, BLEU~\cite{10.3115/1073083.1073135}, Distinct-n~\cite{li-etal-2016-diversity}, and METEOR~\cite{denkowski-lavie-2011-meteor}. For example, ROUGE is the dominant metric in the summarization evaluation area. Its variants consider the overlap of unigrams (ROUGE-1) and bigrams (ROUGE-2), among others. BLEU metric is the common practice for the machine translation evaluation area. Although these metrics achieve good correlations (typically large overlaps) with golden references, they are not general enough because a system summary might convey the same meaning while using different surface forms.

\paragraph{Embedding-based Metrics.}
To further improve semantic similarity between a generated text and a reference text, embedding-based metrics are proposed based on the word embeddings (\emph{e.g.}, WMD~\cite{pmlr-v37-kusnerb15}) or sentence embeddings (\emph{e.g.}, BERTScore~\cite{Zhang2020BERTScoreET} and MoverScore~\cite{zhao-etal-2019-moverscore}). These metrics further bridge the gap with human-related judgment while they largely depend on the quality of embeddings, which may limit their potential.

\paragraph{LLM-based Metrics.} With the development of LLMs, some researchers show that LLMs could achieve great correlation with human judgment, \emph{i.e.}, BARTScore~\cite{https://doi.org/10.48550/arxiv.2106.11520}, and GPTScore~\cite{https://doi.org/10.48550/arxiv.2302.04166}. However, ChatGPT, as a more powerful conversational LLM, has not been investigated to evaluate the quality of the NLG model outputs.

\paragraph{Other Metrics.}
In different research fields, there are some paraphraser-based or task-specific metrics. For example, PRISM~\cite{thompson-post-2020-automatic} is proposed to evaluate translation outputs based on the pre-trained paraphrase models. StoryER~\cite{chen-etal-2022-storyer}, a learning metric, mimics human preference when judging a story by three steps: Ranking, Rating, and Reasoning based on a specific story-generation dataset. Besides, a specifically developed metric named PARENT~\cite{dhingra-etal-2019-handling} is designed for the table2text generation. Other statistical indicators, such as omission errors, hallucination errors, addition errors, duplication errors, and extrinsic errors, are also applied in the table2text task. Although these metrics have obtained impressive results, human evaluation is still inevitable in table2text.

\subsection{Research on ChatGPT}
In recent years, from BERT~\cite{devlin-etal-2019-bert} to ChatGPT~\cite{ChatGPT}, a large number of pre-trained language models have been proposed one after another. Both their parameters and ability are gradually increased, facilitating much-advanced techniques. In particular, ChatGPT, which shows us a revolutionary change as an intelligent conversational large language model, sends shock waves through the research community and industries that have continued to reverberate to this day. With the emergence of ChatGPT, there are two growing research interests related to it: (1) leveraging ChatGPT to deal with various NLP tasks and evaluating its performance using traditional task-specific metrics (\emph{i.e.}, evaluation), and (2) using as a metric to evaluate the outputs of other task-specific models (\emph{i.e.}, evaluator)~\cite{kocmi2023large}.

\paragraph{Evaluation.} Generally, the evaluation tasks on ChatGPT can be divided into two categories, \emph{i.e.}, natural language understanding (NLU) and natural language generation (NLG). For NLU tasks, some researchers find that ChatGPT covers almost all NLU tasks (\emph{e.g.}, sentiment analysis, textual similarity and textual entailment) and achieves competitive or even better performance~\cite{qin2023chatgpt,bang2023multitask,https://doi.org/10.48550/arxiv.2302.10198}.
For NLG tasks, machine translation~\cite{jiao2023chatgpt}, summarization~\cite{Yang2023ExploringTL}, query generation~\cite{https://doi.org/10.48550/arxiv.2302.03495}, and radiology report simplification~\cite{https://doi.org/10.48550/arxiv.2212.14882} are involved. Different from them, we regard ChatGPT as a human evaluator to automatically assess the quality of general textual generations rather than using it for solving tasks.

\paragraph{Evaluator.} As an evaluator, there are two studies that evaluate the quality of translation~\cite{kocmi2023large} and human personalities~\cite{https://doi.org/10.48550/arxiv.2303.01248} by prompting ChatGPT. However, in this work, we aim to evaluate the more general textual outputs to further show the ability of ChatGPT as a general NLG metric.

\section{ChatGPT for NLG Evaluation}
In this section, we discuss how to prompt ChatGPT to serve as a reference-free NLG metric (\S~\ref{subsec:3.1}) or a reference-based NLG metric (\S~\ref{subsec:3.2}) to evaluate the generation quality of NLG models. We take the news summarization task as an example, and give the details of the prompt templates.

\subsection{Reference-free Metric}
\label{subsec:3.1}

To evaluate the generation quality of NLG models, we regard ChatGPT as a human evaluator and give it evaluation instruction via different prompts.
Each prompt should specify (1) which NLG task (\emph{e.g.}, summarization) needs to be evaluated and (2) which aspect (\emph{e.g.}, fluency) of the generation result should be assessed currently.

Inspired by~\citet{kocmi2023large}, we utilize the following two prompts: direct assessment (\textbf{DA}) and one-to-five stars ranking (\textbf{star}).

\begin{framed}
\noindent \centerline{(DA Prompt)}

\noindent Score the following \textcolor[RGB]{101,42,150}{\texttt{[task-ins]}} with respect to \textcolor[RGB]{101,42,150}{\texttt{[aspect]}} on a continuous scale from 0 to 100, where a score of zero means ``\textcolor[RGB]{101,42,150}{\texttt{[ant-aspect]}}'' and score of one hundred means ``perfect \textcolor[RGB]{101,42,150}{\texttt{[aspect]}}''. Note that \textcolor[RGB]{101,42,150}{\texttt{[aspect]}} measures \textcolor[RGB]{101,42,150}{\texttt{[aspect-ins]}}.

\vspace{1ex}
\noindent \textcolor[RGB]{101,42,150}{\texttt{[Conditioned Text]}}

\noindent \textcolor[RGB]{101,42,150}{\texttt{[Generated Text]}}

\noindent Scores:
\end{framed}

\vspace{0.5ex}

\begin{framed}
\noindent \centerline{(Star Prompt)}

\noindent Score the following \textcolor[RGB]{101,42,150}{\texttt{[task-ins]}} with respect to \textcolor[RGB]{101,42,150}{\texttt{[aspect]}} with one to five stars, where one star means ``\textcolor[RGB]{101,42,150}{\texttt{[ant-aspect]}}'' and five stars means ``perfect \textcolor[RGB]{101,42,150}{\texttt{[aspect]}}''. Note that \textcolor[RGB]{101,42,150}{\texttt{[aspect]}} measures \textcolor[RGB]{101,42,150}{\texttt{[aspect-ins]}}.

\vspace{1ex}
\noindent \textcolor[RGB]{101,42,150}{\texttt{[Conditioned Text]}}

\noindent \textcolor[RGB]{101,42,150}{\texttt{[Generated Text]}}

\noindent Stars:
\end{framed}

\vspace{0.5ex}

\noindent where \texttt{[task-ins]} and \texttt{[aspect-ins]} are the instructions of the current task and aspect, respectively. \texttt{[aspect]} and \texttt{[ant-aspect]} denote the evaluated aspect and its antonym, respectively. \texttt{[Conditioned Text]} is the input of NLG models while \texttt{[Generated Text]} is the output.
For example, when evaluating news summarization models in terms of fluency, the DA prompt may be like this:
\begin{framed}
\noindent Score the following \textcolor[RGB]{101,42,150}{news summarization given the corresponding news} with respect to \textcolor[RGB]{101,42,150}{fluency} on a continuous scale from 0 to 100, where a score of zero means "\textcolor[RGB]{101,42,150}{disfluency}" and score of one hundred means "\textcolor[RGB]{101,42,150}{perfect fluency}". Note that \textcolor[RGB]{101,42,150}{fluency} measures \textcolor[RGB]{101,42,150}{the quality of individual sentences, are they well-written and grammatically correct. Consider the quality of individual sentences}.

\vspace{1ex}
\noindent \textcolor[RGB]{101,42,150}{News: [a news article]}

\noindent \textcolor[RGB]{101,42,150}{Summary: [one generated summary]}

\noindent Scores:
\end{framed}

\vspace{0.5ex}

In this manner, both the details of the task and the evaluation aspect are given to ChatGPT. Next, ChatGPT will give its judgment (\emph{e.g.}, ``score: 70'') and the corresponding illustrative description (\emph{e.g.}, ``the summary covers the main points of the news, but ...''). A specific example is shown in Figure~\ref{fig:example}. Finally, the numerical scores could be extracted via several simple heuristic rules.

\subsection{Reference-based Metric}
\label{subsec:3.2}

In addition to reference-free metrics, we explicitly mention the golden references in the prompts to make ChatGPT a reference-based NLG metric:

\vspace{0.5ex}

\begin{framed}
\noindent \centerline{(DA Prompt w/ Reference)}

\noindent Score the following \textcolor[RGB]{101,42,150}{\texttt{[task-ins]}} with respect to \textcolor[RGB]{101,42,150}{\texttt{[aspect]}} on a continuous scale from 0 to 100, where a score of zero means ``\textcolor[RGB]{101,42,150}{\texttt{[ant-aspect]}}'' and score of one hundred means ``perfect \textcolor[RGB]{101,42,150}{\texttt{[aspect]}}''. Note that \textcolor[RGB]{101,42,150}{\texttt{[aspect]}} measures \textcolor[RGB]{101,42,150}{\texttt{[aspect-ins]}}.

\vspace{1ex}
\noindent \textcolor[RGB]{101,42,150}{\texttt{[Conditioned Text]}}

\noindent Human reference: \textcolor[RGB]{101,42,150}{\texttt{[A Reference]}}

\noindent \textcolor[RGB]{101,42,150}{\texttt{[Generated Text]}}

\noindent Scores:
\end{framed}

\vspace{0.5ex}

The star prompt with reference is also formed in a similar way:

\vspace{0.5ex}

\begin{framed}
\noindent \centerline{(Star Prompt w/ Reference))}

\noindent Score the following \textcolor[RGB]{101,42,150}{\texttt{[task-ins]}} with respect to \textcolor[RGB]{101,42,150}{\texttt{[aspect]}} with one to five stars, where one star means ``\textcolor[RGB]{101,42,150}{\texttt{[ant-aspect]}}'' and five stars means ``perfect \textcolor[RGB]{101,42,150}{\texttt{[aspect]}}''. Note that \textcolor[RGB]{101,42,150}{\texttt{[aspect]}} measures \textcolor[RGB]{101,42,150}{\texttt{[aspect-ins]}}.

\vspace{1ex}
\noindent \textcolor[RGB]{101,42,150}{\texttt{[Conditioned Text]}}

\noindent Human reference: \textcolor[RGB]{101,42,150}{\texttt{[A Reference]}}

\noindent \textcolor[RGB]{101,42,150}{\texttt{[Generated Text]}}

\noindent Stars:
\end{framed}

\vspace{0.5ex}

In this way, the ChatGPT evaluator will make its judgment and give the evaluation results under the consideration of the golden references.

\begin{table*}[t]
\centering
\resizebox{1.00\textwidth}{!}
{
\begin{tabular}{lccccccccccccccc}
\toprule[1pt]
\multicolumn{1}{c}{\multirow{2}{*}{Metrics}} & \multicolumn{3}{c}{Coherence}                                               & \multicolumn{3}{c}{Relevance}                    & \multicolumn{3}{c}{Consistency}                  & \multicolumn{3}{c}{Fluency}      & \multicolumn{3}{c}{Avg.}                   \\
\multicolumn{1}{c}{}                         & \multicolumn{1}{l}{Spear.} & \multicolumn{1}{l}{Pear.} & Kend.        & Spear.       & Pear.        & Kend.        & Spear.       & Pear.        & Kend.        & Spear.       & Pear.        & Kend.     & Spear.       & Pear.        & Kend.       \\ \midrule[1pt]
ROUGE-1                                      & 0.167                        & 0.160                       & 0.126          & 0.326          & 0.359          & 0.252          & 0.160          & 0.224          & 0.130          & 0.115          & 0.158          & 0.094    & \cellcolor{ggreen!30}{0.192}  & \cellcolor{ggreen!30}{0.225}  &  \cellcolor{ggreen!30}{0.150}      \\
ROUGE-2                                      & 0.184                        & 0.174                       & 0.139          & 0.290          & 0.327          & 0.219          & 0.187          & 0.246          & 0.155          & 0.159          & 0.185          & 0.128    & \cellcolor{ggreen!30}{0.205}  &  \cellcolor{ggreen!30}{0.233} &  \cellcolor{ggreen!30}{0.160}       \\
ROUGE-L                                      & 0.128                        & 0.102                       & 0.099          & 0.311          & 0.342          & 0.237          & 0.115          & 0.189          & 0.092          & 0.105          & 0.141          & 0.084      &  \cellcolor{ggreen!30}{0.165} & \cellcolor{ggreen!30}{0.194}  &  \cellcolor{ggreen!30}{0.128}     \\
BERTScore                                    & 0.283                        & 0.310                       & 0.211          & 0.311          & 0.346          & 0.243          & 0.110          & 0.152          & 0.090          & 0.192          & 0.209          & 0.158       &  \cellcolor{ggreen!30}{0.224} &  \cellcolor{ggreen!30}{0.254} &  \cellcolor{ggreen!30}{0.175}    \\
MoverScore                                   & 0.159                        & 0.167                       & 0.118          & 0.318          & 0.371          & 0.244          & 0.157          & 0.224          & 0.127          & 0.129          & 0.176          & 0.105       &  \cellcolor{ggreen!30}{0.191}  &  \cellcolor{ggreen!30}{0.234} &  \cellcolor{ggreen!30}{0.148}    \\
PRISM                                        & 0.249                        & 0.258                       & 0.196          & 0.212          & 0.232          & 0.163          & 0.345          & 0.352          & 0.285          & 0.254          & 0.264          & 0.205      & \cellcolor{ggreen!30}{0.265}  &  \cellcolor{ggreen!30}{0.276} &  \cellcolor{ggreen!30}{0.212}     \\
BARTScore                                    & 0.322                        & 0.345                       & 0.250          & 0.264          & 0.290          & 0.197          & 0.311          & 0.321          & 0.256          & 0.248          & 0.260          & 0.203       &  \cellcolor{ggreen!30}{0.286}  &  \cellcolor{ggreen!30}{0.304} &  \cellcolor{ggreen!30}{0.227}    \\
BARTScore+CNN                                & 0.448                        & 0.458                       & 0.342          & 0.356          & 0.369          & 0.273          & 0.382          & 0.422          & 0.315          & 0.356          & 0.407          & 0.292      &  \cellcolor{ggreen!30}{0.385} &  \cellcolor{ggreen!30}{0.414} &  \cellcolor{ggreen!30}{0.305}     \\
BARTScore+CNN+Para                           & 0.424                        & 0.442                       & 0.325          & 0.313          & 0.364          & 0.241          & 0.401          & 0.487          & 0.332          & 0.378 & \textbf{0.448} & 0.311   &  \cellcolor{ggreen!30}{0.379} & \cellcolor{ggreen!30}{0.435}  &  \cellcolor{ggreen!30}{0.302}        \\ \midrule[1pt]
ChatGPT (DA w/o ref)    & 0.451      & 0.456    &  0.383 & 0.439 & 0.473 & \textbf{0.379} & \textbf{0.432} & 0.512 & \textbf{0.399} & \textbf{0.380}     & 0.443    & \textbf{0.351}  & \cellcolor{ggreen!30}{\textbf{0.425}} & \cellcolor{ggreen!30}{\textbf{0.471}}  &  \cellcolor{ggreen!30}{\textbf{0.378}}  \\
ChatGPT (Stars w/o ref)                                      & 0.470               & 0.484              & 0.403 & 0.428 & 0.454 & 0.374 & 0.419 & 0.517 & 0.389 & 0.353          & 0.415          & 0.329  &   \cellcolor{ggreen!30}{0.417} &  \cellcolor{ggreen!30}{0.468} &  \cellcolor{ggreen!30}{0.374}  \\ 
ChatGPT (DA w/ ref)   & 0.420               & 0.435              & 0.346 & \textbf{0.448} & \textbf{0.474} & 0.378 & 0.424 & \textbf{0.534} & 0.384 & 0.316   & 0.381      &  0.285  & \cellcolor{ggreen!30}{0.402} &  \cellcolor{ggreen!30}{0.456} &  \cellcolor{ggreen!30}{0.348}  \\
ChatGPT (Stars w/ ref)  & \textbf{0.474}    & \textbf{0.491}      & \textbf{0.407} & 0.430 & 0.457 & 0.378 &  0.403 & 0.489 & 0.375 & 0.339     & 0.409    & 0.319   & \cellcolor{ggreen!30}{0.411}  & \cellcolor{ggreen!30}{0.461}  &  \cellcolor{ggreen!30}{0.370} \\   \bottomrule[1pt]
\end{tabular}
}
\caption{Sample-level Spearman correlation (Spear.) correlation, Pearson (Pear.) correlation and Kendall's Tau (Kend.) of different aspects on SummEval (a text summarization meta-evaluation dataset). ``\colorbox{ggreen!30}{Avg.}'' indicates the average performance. The \textbf{bold} indicates the best correlation.}
\label{table:sample_level_summeval}
\end{table*}

\section{Experiments}

\subsection{Experimental Setup}

\noindent \textbf{Metrics.} To evaluate how well automatic metrics correlate with human judgment. Two widely-used correlation measures are adopted: (1) \emph{Spearman correlation}~\cite{zar2005spearman} assesses the monotonic relationships between two variables; (2) \emph{Pearson correlation}~\cite{mukaka2012guide} measures the linear relationships between two sets of data; (3) \emph{Kendall's Tau}~\cite{kendall1938new} evaluates the ordinal association between two measured quantities.

\vspace{0.5ex}
\noindent \textbf{Evaluation Strategy.} When calculating the correlation scores, there are different aggregation methods. Given a set of conditioned text $\{c_1, c_2, ..., c_n\}$ (\emph{e.g.}, source documents in text summarization task) and $M$ NLG models. The generated text of $m$-th model for the $i$-th condition text is denoted as $g_{i,m}$. (1) \emph{Sample-level} evaluation strategy calculates the correlation scores as follows:

\begin{equation}
\begin{aligned}
\small
\mathrm{Corr}_\text{sample} =& \frac{1}{n} \sum^{n}_{1} (\rho ([ f_\text{auto}(g_{i,1}), ...,  f_\text{auto}(g_{i,M})], \\
&[ f_\text{human}(g_{i,1}), ...,  f_\text{human}(g_{i,M})]))
\end{aligned}  
\end{equation}
where $\rho$ denotes the correlation metrics like Spearman correlation. $f_\text{auto}$ and $f_\text{human}$ indicate the automatic evaluation and human judgment functions, respectively.

\noindent (2) \emph{Dataset-level} evaluation strategy calculates the correlation as follows:
\begin{equation}
\begin{aligned}
\small
\mathrm{Corr}_\text{dataset} =& \rho ([ f_\text{auto}(g_{1,1}), ...,  f_\text{auto}(g_{n,M})], \\
&[ f_\text{human}(g_{1,1}), ...,  f_\text{human}(g_{n,M})])
\end{aligned}  
\end{equation}

\begin{table*}[t]
\centering
\resizebox{1.00\textwidth}{!}
{
\begin{tabular}{lccccccccccccccc}
\toprule[1pt]
\multicolumn{1}{c}{\multirow{2}{*}{Metrics}} & \multicolumn{3}{c}{Coherence}                                               & \multicolumn{3}{c}{Relevance}                    & \multicolumn{3}{c}{Informativeness}                  & \multicolumn{3}{c}{Fluency}      & \multicolumn{3}{c}{Avg.}                   \\
\multicolumn{1}{c}{}                         & \multicolumn{1}{l}{Spear.} & \multicolumn{1}{l}{Pear.} & Kend.        & Spear.       & Pear.        & Kend.        & Spear.       & Pear.        & Kend.        & Spear.       & Pear.        & Kend.     & Spear.       & Pear.        & Kend.       \\ \midrule[1pt]
ROUGE-1     &  0.095  &     -0.071     &   0.076  &  0.147    &    -0.001    &  0.112 &   0.130  &   -0.039       &   0.099     &   0.104      &  -0.074    &  0.082   & \cellcolor{ggreen!30}{0.119}  & \cellcolor{ggreen!30}{-0.046}  &  \cellcolor{ggreen!30}{0.092}      \\
ROUGE-2     &  0.026  &   -0.108       & 0.009    &   0.091   &   -0.056     & 0.065  & 0.079   &    -0.087       &   0.052     &    0.048     &   -0.101   & 0.032    & \cellcolor{ggreen!30}{0.061}  & \cellcolor{ggreen!30}{-0.088}  &  \cellcolor{ggreen!30}{0.092}      \\
ROUGE-L    & 0.064   &   -0.091       &  0.051   &   0.106   &   -0.034     & 0.083  &  0.089  &    -0.068       &    0.064    &   0.072      &  -0.090    &  0.061   & \cellcolor{ggreen!30}{0.083}  & \cellcolor{ggreen!30}{-0.071}  &  \cellcolor{ggreen!30}{0.065}      \\
BERTScore  &  0.147  &   0.043       &  0.116   &   0.162   &    0.080    &  0.126 &  0.130  &     0.044      &  0.105      &   0.171      &   0.046   &  0.128   & \cellcolor{ggreen!30}{0.152}  & \cellcolor{ggreen!30}{0.053}  &  \cellcolor{ggreen!30}{0.119}      \\
MoverScore  &  0.161  &   0.008       &  0.127   &  0.195    &   0.077     & 0.157  & 0.188   &    0.045       &  0.151      &   0.120      &   -0.008   &  0.086   & \cellcolor{ggreen!30}{0.166}  & \cellcolor{ggreen!30}{0.030}  &  \cellcolor{ggreen!30}{0.130}      \\
PRISM    &  0.573  &    0.605      &  0.478   &  0.553    &    0.636    & 0.460  &  0.561  &    0.629       &    0.472    &   0.532      &  0.547    &  0.443   & \cellcolor{ggreen!30}{0.555}  & \cellcolor{ggreen!30}{0.604}  &  \cellcolor{ggreen!30}{0.463}      \\
BARTScore    &  \textbf{0.679}  &   \textbf{0.709}       & \textbf{0.568}    &  \textbf{0.604}    &  \textbf{0.744}      & \textbf{0.507}  &  \textbf{0.646}  &    \textbf{0.749}       &    \textbf{0.543}    &   \textbf{0.670}      &   \textbf{0.662}   &   \textbf{0.564}  & \cellcolor{ggreen!30}{\textbf{0.650}}  & \cellcolor{ggreen!30}{\textbf{0.716}}  &  \cellcolor{ggreen!30}{\textbf{0.545}}      \\
BARTScore+CNN    &  0.653  &      0.690    &   0.547  &   0.567   &  0.718      &  0.478 & 0.616   &   0.712       &     0.510   &   0.640      &  0.653    &  0.540   & \cellcolor{ggreen!30}{0.619}  & \cellcolor{ggreen!30}{0.693}  &  \cellcolor{ggreen!30}{0.519}      \\
BARTScore+CNN+Para  &  0.657  &   0.675       &   0.544  &  0.562    &  0.739      & 0.465  &  0.614  &    0.727       &  0.507      &   0.652      &   0.630   &   0.545  & \cellcolor{ggreen!30}{0.621}  & \cellcolor{ggreen!30}{0.693}  &  \cellcolor{ggreen!30}{0.515}      \\ \midrule[1pt]
ChatGPT (DA w/o ref)    &  0.469  &      0.487    &  0.405   &  0.461    &  0.587      & 0.392  &  0.578  &   0.645        &   0.498     &    0.507     &   0.524   &  0.427   & \cellcolor{ggreen!30}{0.504}  & \cellcolor{ggreen!30}{0.561}  &  \cellcolor{ggreen!30}{0.430}      \\
ChatGPT (Stars w/o ref)     &  0.428  &     0.424     &  0.375   &  0.402    &   0.488     & 0.348  &  0.557  &     0.602      &  0.487      &   0.451     &   0.448   &   0.385  & \cellcolor{ggreen!30}{0.460}  & \cellcolor{ggreen!30}{0.490}  &  \cellcolor{ggreen!30}{0.399}      \\
ChatGPT (DA w/ ref)   &  0.431  &    0.494      &  0.369   &   0.436   &   0.535     & 0.372  &  0.429  &    0.484       &  0.368      &     0.459    &  0.490    &  0.387   & \cellcolor{ggreen!30}{0.439}  & \cellcolor{ggreen!30}{0.501}  &  \cellcolor{ggreen!30}{0.374}      \\
ChatGPT (Stars w/ ref) &  0.423  &    0.424      &  0.369    &   0.443   &    0.506    & 0.395   &  0.404  &    0.463       &    0.352    &     0.503    &  0.504    &  0.430   & \cellcolor{ggreen!30}{0.443}  & \cellcolor{ggreen!30}{0.474}  &  \cellcolor{ggreen!30}{0.387}      \\  \bottomrule[1pt]
\end{tabular}
}
\caption{Sample-level Spearman correlation (Spear.) correlation, Pearson (Pear.) correlation and Kendall's Tau (Kend.) of different aspects on NewsRoom (a text summarization meta-evaluation dataset). ``\colorbox{ggreen!30}{Avg.}'' indicates the average performance. The \textbf{bold} indicates the best correlation.}
\label{table:sample_level_newsroom}
\end{table*}

\vspace{1.0ex}

\begin{table}[t]
\centering
\resizebox{0.50\textwidth}{!}
{
\begin{tabular}{lcccccc}
\toprule[1pt]
\multicolumn{1}{c}{\multirow{2}{*}{Metrics}} & \multicolumn{3}{c}{Sample-level}                 & \multicolumn{3}{c}{Dataset-level}                \\
\multicolumn{1}{c}{}                         & Spear.       & Pear.        & Kend.        & Spear.       & Pear.        & Kend.        \\ \midrule[1pt]
ROUGE-1     &   \textbf{0.498}    &   \textbf{0.526}     &  \textbf{0.408}     &  \textbf{0.533}    &    \textbf{0.555}     & \textbf{0.383}        \\
ROUGE-2    &   0.423    &   0.449     &  0.353     &   0.514   &    0.513     &    0.369      \\
ROUGE-L   &  0.488     &   0.515     &    0.403   &    \textbf{0.533}  &    0.549     &  \textbf{0.383}        \\
BERTScore    &  0.441     &   0.477     &   0.347    &   0.503   &    0.517     &  0.358        \\
MoverScore    &   0.372    &   0.400     &   0.290    &  0.427    &    0.451     &    0.303      \\
PRISM     &   0.411    &   0.458     &   0.324    &   0.478   &   0.494   &  0.339        \\
BARTScore     &   0.441    &   0.467     &  0.342     &    0.467  &  0.458       &   0.327       \\
BARTScore+CNN     &  0.475     & 0.500      &   0.374    &  0.436    &   0.455      &   0.306       \\
BARTScore+CNN+Para         &   0.471    &   0.512     &   0.374    &   0.499   &   0.515      &  0.357    \\ \midrule[1pt]
ChatGPT (DA w/o ref)    &    0.173   &    0.179    &   0.152    &   0.185   &     0.193    &   0.145       \\
ChatGPT (Stars w/o ref)    &  0.145     &   0.162     &  0.129    &   0.170   &  0.179       &    0.136      \\
ChatGPT (DA w/ ref)    &   0.184    &   0.208     &   0.154    &    0.276  &  0.288       &  0.206        \\
ChatGPT (Stars w/ ref)    &   0.195    &   0.207     &   0.174    &   0.224   &   0.261      &   0.181       \\ \bottomrule[1pt]
\end{tabular}
}
\caption{Sample-level and Dataset-level correlation on RealSumm (a text summarization meta-evaluation dataset) (Spear.: Spearman correlation; Pear.: Pearson correlation; Kend.: Kendall's Tau). The \textbf{bold} indicates the best correlation.}
\label{table:all_level_realsumm}
\end{table}

\subsection{Baselines}
We compare the ChatGPT evaluator with the following widely-used automatic NLG metrics to provide deeper analyses:

\begin{itemize}[leftmargin=*,topsep=0pt]
\setlength{\itemsep}{0pt}
\setlength{\parsep}{0pt}
\setlength{\parskip}{0pt}
\item ROUGE-1, ROUGE-2 and ROUGE-L~\cite{lin-2004-rouge} measure the lexical overlap between the generated text and corresponding references based on unigram, bigram and longest common subsequence, respectively.
\vspace{0.5ex}
\item BERTScore~\cite{Zhang2020BERTScoreET} and MoverScore~\cite{zhao-etal-2019-moverscore} evaluate the semantic similarity via pre-trained BERT model~\cite{devlin-etal-2019-bert}.
\vspace{0.5ex}
\item PRISM~\cite{thompson-post-2020-automatic} is used to evaluate NLG models via pre-trained paraphrase models.
\vspace{0.5ex}
\item BARTScore~\cite{https://doi.org/10.48550/arxiv.2106.11520} is a state-of-the-art NLG metrics based on vanilla pre-trained BART model~\cite{lewis-etal-2020-bart}.
\vspace{0.5ex}
\item BARTScore+CNN~\cite{https://doi.org/10.48550/arxiv.2106.11520} could be regarded as an enhanced version of BARTScore. This metric is based on the BART fine-tuned on the CNN/DM dataset~\cite{Hermann2015TeachingMT}.
\vspace{0.5ex}
\item BARTScore+CNN+Para~\cite{https://doi.org/10.48550/arxiv.2106.11520} is another enhanced version of BARTScore. The metric is based on the BART fine-tuned on both CNN/DM and Paraphrase2.0~\cite{hu-etal-2019-large}.
\vspace{0.5ex}
\item Perplexity (PPL) is a commonly-used NLG metric to evaluate whether the generation result is grammatical and fluent.
\end{itemize}

\begin{table*}[t]
\centering
\resizebox{1.00\textwidth}{!}
{
\begin{tabular}{lccccccccccccccc}
\toprule[1pt]
\multicolumn{1}{c}{\multirow{2}{*}{Metrics}} & \multicolumn{3}{c}{Coherence}                    & \multicolumn{3}{c}{Relevance}                    & \multicolumn{3}{c}{Consistency}                  & \multicolumn{3}{c}{Fluency}           & \multicolumn{3}{c}{Avg.}              \\
\multicolumn{1}{c}{}                         & Spear.         & Pear.          & Kend.          & Spear.         & Pear.          & Kend.          & Spear.         & Pear.          & Kend.          & Spear.         & Pear.          & Kend.     & Spear.         & Pear.          & Kend.        \\ \midrule[1pt]
ROUGE-1                                      & 0.184          & 0.193          & 0.129          & 0.302          & 0.341          & 0.217          & 0.137          & 0.175          & 0.108          & 0.080          & 0.143          & 0.062    & \cellcolor{ggreen!30}{0.176} &  \cellcolor{ggreen!30}{0.213}  &  \cellcolor{ggreen!30}{0.129}     \\
ROUGE-2                                      & 0.145          & 0.140          & 0.102          & 0.245          & 0.254          & 0.175          & 0.129          & 0.152          & 0.102          & 0.062          & 0.092          & 0.048      &  \cellcolor{ggreen!30}{0.145} &  \cellcolor{ggreen!30}{0.160} &  \cellcolor{ggreen!30}{0.107}     \\
ROUGE-L                                      & 0.141          & 0.148          & 0.100          & 0.284          & 0.318          & 0.204          & 0.109          & 0.152          & 0.086          & 0.079          & 0.132          & 0.061      &  \cellcolor{ggreen!30}{0.153} & \cellcolor{ggreen!30}{0.188} & \cellcolor{ggreen!30}{0.113}     \\
BERTScore                                    & 0.317          & 0.326          & 0.224          & 0.362          & 0.381          & 0.262          & 0.117          & 0.146          & 0.092          & 0.150          & 0.196          & 0.117       &  \cellcolor{ggreen!30}{0.237} & \cellcolor{ggreen!30}{0.262} &  \cellcolor{ggreen!30}{0.174}    \\
MoverScore                                   & 0.178          & 0.177          & 0.125          & 0.294          & 0.328          & 0.211          & 0.150          & 0.171          & 0.118          & 0.119          & 0.160          & 0.092        &  \cellcolor{ggreen!30}{0.185}  & \cellcolor{ggreen!30}{0.209}  &  \cellcolor{ggreen!30}{0.136}   \\
PRISM                                        & 0.286          & 0.281          & 0.204          & 0.280          & 0.297          & 0.202          & 0.323          & 0.297          & 0.256          & 0.236          & 0.248          & 0.184        & \cellcolor{ggreen!30}{0.281}  & \cellcolor{ggreen!30}{0.281} &  \cellcolor{ggreen!30}{0.212}   \\
BARTScore                                    & 0.335          & 0.357          & 0.241          & 0.363          & 0.386          & 0.263          & 0.269          & 0.276          & 0.212          & 0.187          & 0.206          & 0.146       & \cellcolor{ggreen!30}{0.288} &  \cellcolor{ggreen!30}{0.306} &  \cellcolor{ggreen!30}{0.215}    \\
BARTScore+CNN                                & 0.408          & 0.434          & 0.292          & 0.394          & 0.423          & 0.286          & 0.334          & 0.377          & 0.264          & 0.285          & 0.354          & 0.223        & \cellcolor{ggreen!30}{0.355} & \cellcolor{ggreen!30}{0.397} & \cellcolor{ggreen!30}{0.266}    \\
BARTScore+CNN+Para                           & 0.424          & 0.430          & 0.304          & 0.398          & 0.431          & 0.289          & \textbf{0.379}         & 0.452          & 0.301          & \textbf{0.346} & \textbf{0.410} & \textbf{0.271}  & \cellcolor{ggreen!30}{\textbf{0.387}} &  \cellcolor{ggreen!30}{0.431} &  \cellcolor{ggreen!30}{0.291} \\ \midrule[1pt]
ChatGPT (DA w/o ref)    &  0.394    &  0.399   & 0.310  &  0.455  &  0.435  & 0.365     &  0.339   &  0.500   &  0.300   &   0.286   & 0.380      &  0.250   & \cellcolor{ggreen!30}{0.368} & \cellcolor{ggreen!30}{0.428} & \cellcolor{ggreen!30}{0.306}   \\
ChatGPT (Stars w/o ref)                                     & 0.435 & 0.438 & 0.353 & 0.448 & 0.459 & 0.366 & 0.356 & \textbf{0.515} & \textbf{0.320} & 0.300          & 0.385          & 0.268        & \cellcolor{ggreen!30}{0.385} & \cellcolor{ggreen!30}{\textbf{0.449}} &  \cellcolor{ggreen!30}{0.327}   \\ 
ChatGPT (DA w/ ref)    & 0.418    & 0.426    &  0.327  & \textbf{0.494}   &  \textbf{0.506}  &  \textbf{0.389}    &  0.363   &  0.507   & 0.315    &  0.237    &   0.329   & 0.203   & \cellcolor{ggreen!30}{0.378} &  \cellcolor{ggreen!30}{0.442} & \cellcolor{ggreen!30}{0.308}   \\
ChatGPT (Stars w/ ref)    &   \textbf{0.465}  & \textbf{0.472}    & \textbf{0.385}   & 0.458   &  0.476  &  0.381    &  0.333   & 0.475    & 0.299    &   0.285   &  0.385     &  0.258   & \cellcolor{ggreen!30}{0.385} & \cellcolor{ggreen!30}{0.452} & \cellcolor{ggreen!30}{\textbf{0.331}}   \\\bottomrule[1pt]
\end{tabular}
}
\caption{Dataset-level Spearman correlation (Spear.) correlation, Pearson (Pear.) correlation and Kendall's Tau (Kend.) of different aspects on SummEval (a text summarization meta-evaluation dataset). ``\colorbox{ggreen!30}{Avg.}'' indicates the average performance. The \textbf{bold} indicates the best correlation.}
\label{table:dataset_level_summeval}
\end{table*}

\vspace{1.0ex}

\begin{table*}[t]
\centering
\resizebox{1.00\textwidth}{!}
{
\begin{tabular}{lccccccccccccccc}
\toprule[1pt]
\multicolumn{1}{c}{\multirow{2}{*}{Metrics}} & \multicolumn{3}{c}{Coherence}                                               & \multicolumn{3}{c}{Relevance}                    & \multicolumn{3}{c}{Informativeness}                  & \multicolumn{3}{c}{Fluency}      & \multicolumn{3}{c}{Avg.}                   \\
\multicolumn{1}{c}{}                         & \multicolumn{1}{l}{Spear.} & \multicolumn{1}{l}{Pear.} & Kend.        & Spear.       & Pear.        & Kend.        & Spear.       & Pear.        & Kend.        & Spear.       & Pear.        & Kend.     & Spear.       & Pear.        & Kend.       \\ \midrule[1pt]
ROUGE-1     &  0.100  &    0.015      &   0.070  &   0.122   &  0.061      &  0.084 &  0.149  &   0.043    &  0.106      &    0.064     &  -0.009    &  0.043   & \cellcolor{ggreen!30}{0.109}  & \cellcolor{ggreen!30}{0.028}  &  \cellcolor{ggreen!30}{0.076}      \\
ROUGE-2     &  0.080  &    0.033      &  0.060   &  0.124    &   0.071     & 0.092  &  0.158  &  0.060     &    0.119    &   0.045      &  0.018    &   0.032  & \cellcolor{ggreen!30}{0.102}  & \cellcolor{ggreen!30}{0.045}  &  \cellcolor{ggreen!30}{0.076}      \\
ROUGE-L    &  0.079  &    -0.010      &  0.055   &  0.101    &   0.031     & 0.069  &  0.136  &    0.018    &   0.098     &   0.045      &  -0.030    &  0.029   & \cellcolor{ggreen!30}{0.090}  & \cellcolor{ggreen!30}{0.002}  &  \cellcolor{ggreen!30}{0.063}      \\
BERTScore  &  0.169   &   0.138       &   0.122  &   0.176   &   0.158     & 0.127  &  0.196  &  0.153   &  0.141      &  0.154       &   0.129   &   0.109  & \cellcolor{ggreen!30}{0.174}  & \cellcolor{ggreen!30}{0.145}  &  \cellcolor{ggreen!30}{0.125}      \\
MoverScore  &  0.173  &     0.119     &  0.122   &  0.192    &   0.156     &  0.132 &  0.232   &   0.148  &  0.166      &     0.112    &  0.091    &  0.076   & \cellcolor{ggreen!30}{0.177}  & \cellcolor{ggreen!30}{0.129}  &  \cellcolor{ggreen!30}{0.124}      \\
PRISM    &  0.483  &   0.485       &  0.350   &   0.540   &  0.550      & 0.398  &  0.567  &   0.569   &  0.414      &  0.420       &   0.421   &   0.303  & \cellcolor{ggreen!30}{0.503}  & \cellcolor{ggreen!30}{0.506}  &  \cellcolor{ggreen!30}{0.366}      \\
BARTScore    &  \textbf{0.656}  &     \textbf{0.666}     &  \textbf{0.495}   &  \textbf{0.588}    &    \textbf{0.700}    & \textbf{0.439}  & \textbf{0.645}   &      \textbf{0.710}     &     \textbf{0.485}   &   \textbf{0.615}      &   \textbf{0.610}   &   \textbf{0.464}  & \cellcolor{ggreen!30}{\textbf{0.626}}  & \cellcolor{ggreen!30}{\textbf{0.671}}  &  \cellcolor{ggreen!30}{\textbf{0.471}}      \\
BARTScore+CNN    &  0.623  &    0.640      &   0.466  & 0.557     &   0.665     &  0.411 &  0.592  &     0.665      &   0.440     &     0.596    &   0.592   &   0.448  & \cellcolor{ggreen!30}{0.592}  & \cellcolor{ggreen!30}{0.641}  &  \cellcolor{ggreen!30}{0.441}      \\
BARTScore+CNN+Para  &   0.621  &     0.639     &  0.465   &    0.575  &    0.692    & 0.427  &  0.615  &     0.694      &     0.459   &     0.593    &  0.577    &   0.444  & \cellcolor{ggreen!30}{0.601}  & \cellcolor{ggreen!30}{0.650}  &  \cellcolor{ggreen!30}{0.449}      \\ \midrule[1pt]
ChatGPT (DA w/o ref)    &  0.383  &    0.418      &  0.297   &    0.491  &  0.541      &  0.392 &  0.527  &    0.576       &    0.413    &     0.401    &   0.398   &  0.309   & \cellcolor{ggreen!30}{0.451}  & \cellcolor{ggreen!30}{0.483}  &  \cellcolor{ggreen!30}{0.353}      \\
ChatGPT (Stars w/o ref)     &  0.370  &     0.374     &  0.294   &   0.422   &   0.444     & 0.343  & 0.518   & 0.527          &   0.423     &  0.381       &   0.362   &  0.302   & \cellcolor{ggreen!30}{0.423}  & \cellcolor{ggreen!30}{0.427}  &  \cellcolor{ggreen!30}{0.341}      \\
ChatGPT (DA w/ ref)   &  0.381  &     0.407     &  0.292   &   0.434   &    0.458    &  0.339 &  0.377  &   0.412    &  0.291      &   0.386      &  0.403    &  0.298   & \cellcolor{ggreen!30}{0.394}  & \cellcolor{ggreen!30}{0.420}  &  \cellcolor{ggreen!30}{0.305}      \\
ChatGPT (Stars w/ ref) &  0.370  &     0.355     &  0.295   &  0.425    &    0.426    & 0.342  &  0.373  &    0.400       &  0.301      &   0.439      &  0.425    &  0.353   & \cellcolor{ggreen!30}{0.402}  & \cellcolor{ggreen!30}{0.402}  &  \cellcolor{ggreen!30}{0.323}      \\  \bottomrule[1pt]
\end{tabular}
}
\caption{Dataset-level Spearman correlation (Spear.) correlation, Pearson (Pear.) correlation and Kendall's Tau (Kend.) of different aspects on NewsRoom (a text summarization meta-evaluation dataset). ``\colorbox{ggreen!30}{Avg.}'' indicates the average performance. The \textbf{bold} indicates the best correlation.}
\label{table:dataset_level_newsroom}
\end{table*}

\subsection{Text Summarization}
\label{subsec:4.3}

We conduct meta-evaluation on SummEval~\cite{fabbri-etal-2021-summeval}, NewsRoom~\cite{grusky-etal-2018-newsroom} and RealSumm~\cite{bhandari-etal-2020-evaluating} to evaluate the performance of ChatGPT as an NLG metric for text summarization.
SummEval collects 16 model-generated summaries on the CNN/DM dataset and annotates human judgments upon these summaries covering aspects of coherence, relevance, consistency and fluency.
Newsroom, as a text summarization dataset, also provides human judgments on 7 model-generated summaries, including coherence, relevance, informativeness and fluency.
RealSumm evaluates the pyramid~\cite{nenkova-passonneau-2004-evaluating} recall of 25 model-generated summaries.

\vspace{1.0ex}
\noindent \textbf{The Potentiality of ChatGPT.}
Table~\ref{table:sample_level_summeval} and Table~\ref{table:sample_level_newsroom} show the sample-level evaluation results on SummEval and NewsRoom, respectively (dataset-level evaluation results on SummEval and NewsRoom also shown in Table~\ref{table:dataset_level_summeval} and Table~\ref{table:dataset_level_newsroom} with the similar trends).
Experimental results show that ChatGPT achieves a new state-of-the-art correlation in most aspects of SummEval, demonstrating its potential of serving as an NLG metric. For results on Newsroom, ChatGPT also outperforms dominant summarization metrics (\emph{i.e.}, ROUGE and BERTScore) by a large margin.
Note that our experiments only estimate the lower bound of ChatGPT's performance, and better performances would like to be achieved once using better prompts or updated versions of ChatGPT.

\begin{table}[t]
\centering
\resizebox{0.50\textwidth}{!}
{
\begin{tabular}{lcccccc}
\toprule[1pt]
\multicolumn{1}{c}{\multirow{2}{*}{Metrics}} & \multicolumn{3}{c}{Sample-level}                 & \multicolumn{3}{c}{Dataset-level}                \\
\multicolumn{1}{c}{}                         & Spear.       & Pear.        & Kend.        & Spear.       & Pear.        & Kend.        \\ \midrule[1pt]
ROUGE-1                                      & 0.014          & 0.020          & 0.013          & -0.023         & -0.010         & -0.016         \\
ROUGE-2                                      & 0.035          & 0.041          & 0.035          & 0.009          & 0.012          & 0.007          \\
ROUGE-L                                      & 0.013          & 0.021          & 0.014          & -0.016         & -0.004         & -0.011         \\
BERTScore                                    & 0.140          & 0.120          & 0.105          & 0.081          & 0.084          & 0.056          \\
BARTScore                                    & -0.065         & -0.082         & -0.061         & -0.065         & -0.092         & -0.045         \\
BARTScore+CNN                                & 0.049          & 0.026          & 0.041          & 0.047          & 0.053          & 0.033          \\
BARTScore+CNN+Para                           & 0.064          & 0.050          & 0.062          & 0.062          & 0.074          & 0.043          \\
PPL                                          & 0.324          & 0.330          & 0.265          & 0.306          & 0.255          & 0.213          \\ \midrule[1pt]
ChatGPT (DA w/o ref) & \textbf{0.507} &  \textbf{0.533} &  \textbf{0.439} & \textbf{0.471} & \textbf{0.494} & \textbf{0.366} \\ 
ChatGPT (Stars w/o ref)                                      & 0.472 & 0.490 & 0.427 & 0.415 & 0.439 & 0.342 \\
ChatGPT (DA w/ ref) & 0.411 & 0.434 & 0.357 & 0.363 & 0.375 & 0.281 \\ 
ChatGPT (Stars w/ ref) & 0.478 & 0.493 & 0.435 & 0.346 & 0.372 & 0.291 \\ \bottomrule[1pt]
\end{tabular}
}
\caption{Sample-level and Dataset-level correlation on OpenMEVA (a story generation meta-evaluation dataset) (Spear.: Spearman correlation; Pear.: Pearson correlation; Kend.: Kendall's Tau).}
\label{table:all_level_openmeva}
\end{table}

\begin{table*}[t]
\centering
\resizebox{1.00\textwidth}{!}
{
\begin{tabular}{lcccccccccccc}
\toprule[1pt]
\multicolumn{1}{c}{\multirow{2}{*}{Metrics}} & \multicolumn{3}{c}{Informativeness}              & \multicolumn{3}{c}{Naturalness}                  & \multicolumn{3}{c}{Quality} & \multicolumn{3}{c}{Avg.}  \\
\multicolumn{1}{c}{}                         & Spear.         & Pear.          & Kend.          & Spear.         & Pear.          & Kend.          & Spear.         & Pear.          & Kend.   & Spear.         & Pear.          & Kend.       \\ \midrule[1pt]
ROUGE-1                                      & 0.092          & 0.093          & 0.073          & 0.265          & 0.274          & 0.206          & 0.235          &    0.234       & 0.184  & \cellcolor{ggreen!30}{0.197} & \cellcolor{ggreen!30}{0.200} & \cellcolor{ggreen!30}{0.154}        \\
ROUGE-2                                      & 0.133          & 0.137          & 0.103          & 0.233          & 0.241          & 0.177          & 0.192          & 0.192          &  0.145  & \cellcolor{ggreen!30}{0.186} & \cellcolor{ggreen!30}{0.190} & \cellcolor{ggreen!30}{0.142}       \\
ROUGE-L                                      & 0.079          & 0.084          & 0.063          & 0.237          & 0.255          & 0.183          & 0.210          &     0.216      & 0.163  & \cellcolor{ggreen!30}{0.175} & \cellcolor{ggreen!30}{0.185} & \cellcolor{ggreen!30}{0.136}        \\
BERTScore                                    & 0.231          & 0.261          & 0.174          & 0.288          & 0.327          & 0.216          & 0.264          &  0.304          &  0.197  & \cellcolor{ggreen!30}{0.261} & \cellcolor{ggreen!30}{0.297} & \cellcolor{ggreen!30}{0.196}       \\
MoverScore                                   & \textbf{0.284} & \textbf{0.276} & 0.209          & 0.189          & 0.189          & 0.140          & 0.152          &  0.161          &  0.114   & \cellcolor{ggreen!30}{0.208} & \cellcolor{ggreen!30}{0.209} & \cellcolor{ggreen!30}{0.154}       \\
PRISM                                        & 0.255          & 0.268          & 0.189          & 0.301          & 0.350          & 0.223          & 0.308          & 0.337          &   0.226    & \cellcolor{ggreen!30}{0.288} & \cellcolor{ggreen!30}{0.318} & \cellcolor{ggreen!30}{0.213}    \\
BARTScore                                    & 0.234          & 0.270          & 0.174          & 0.221          & 0.280          & 0.164          & 0.186          & 0.245          &  0.139  & \cellcolor{ggreen!30}{0.214} & \cellcolor{ggreen!30}{0.265} & \cellcolor{ggreen!30}{0.159}       \\
BARTScore+CNN                                & 0.237          & 0.253          & 0.177          & 0.312          & 0.382          & 0.233          & 0.294          & 0.357          & 0.219    & \cellcolor{ggreen!30}{0.281} & \cellcolor{ggreen!30}{0.331} & \cellcolor{ggreen!30}{0.210}       \\
BARTScore+CNN+Para                           & 0.240          & 0.266          & 0.177          & \textbf{0.335} & \textbf{0.416} & 0.248          & \textbf{0.343} & \textbf{0.383}         &  0.255  & \cellcolor{ggreen!30}{\textbf{0.306}} & \cellcolor{ggreen!30}{\textbf{0.355}} & \cellcolor{ggreen!30}{0.227}\\ \midrule[1pt]
ChatGPT (DA w/o ref) & - & - & -  & 0.243 & 0.293 & 0.202 & 0.310 & 0.319 & 0.260  & \cellcolor{ggreen!30}{-} & \cellcolor{ggreen!30}{-} & \cellcolor{ggreen!30}{-} \\
 ChatGPT (Stars w/o ref)                                     & -          & -          & - & 0.316          & 0.389          & \textbf{0.269} & 0.307          & 0.367 & \textbf{0.266}  & \cellcolor{ggreen!30}{-} & \cellcolor{ggreen!30}{-} & \cellcolor{ggreen!30}{-}       \\ 
 ChatGPT (DA w/ ref) & 0.247 & 0.255 & 0.198   & 0.305  & 0.344  & 0.248   & 0.269 & 0.343 & 0.215 & \cellcolor{ggreen!30}{0.274} & \cellcolor{ggreen!30}{0.314} & \cellcolor{ggreen!30}{0.220} \\
 ChatGPT (Stars w/ ref) & 0.266 & 0.262 & \textbf{0.224} & 0.293 & 0.374 & 0.235 & 0.276 & 0.328 &  0.239 & \cellcolor{ggreen!30}{0.278} & \cellcolor{ggreen!30}{0.321} & \cellcolor{ggreen!30}{\textbf{0.233}}\\\bottomrule[1pt]
\end{tabular}
}
\caption{Dataset-level Spearman correlation (Spear.) correlation, Pearson (Pear.) correlation and Kendall's Tau (Kend.) of different aspects on BAGEL (a data-to-text generation meta-evaluation dataset). ``\colorbox{ggreen!30}{Avg.}'' indicates the average performance. The \textbf{bold} indicates the best correlation.}
\label{table:dataset_level_bagel}
\end{table*}

\vspace{1.0ex}
\noindent \textbf{The Impact of Dataset Biases.}
As shown in Table~\ref{table:all_level_realsumm}, we find that the experimental results on RealSumm show different trends from those on SummEval, \emph{i.e.}, ChatGPT significantly underperforms other baseline metrics.
For example, ChatGPT (Stars w/ ref) achieves 0.195 sample-level Spearman correlation, which is far behind the counterpart of ROUGE-1 (\emph{i.e.}, 0.498).
We conjecture this is because the human judgments in RealSumm are collected via pyramid method~\cite{nenkova-passonneau-2004-evaluating}. In detail, this method first requires human evaluators to extract semantic content units from golden references, and then score each system summary based on how many extracted semantic content units are mentioned in the system summary.

In this manner, the more similarity between one generated summary and the corresponding golden reference, the more human evaluation scores will be achieved. Therefore, this reference-oriented annotation method makes the traditional $n$-gram-based metric (such as ROUGE) already achieve well correlations with human judgments, which we named as \emph{lexical biases.}
As for SummEval and NewsRoom, human evaluators are required to directly score different summaries without comparing them with the golden references, and thus do not involve such lexical biases.

\vspace{1.0ex}
\noindent \textbf{The Impact of Different Prompt.}
In this work, we attempt four prompts to guide ChatGPT to evaluate the generation of NLG models.
As we can see, the performances of ChatGPT are sensitive to the prompt design. For different aspects, the prompt should be carefully designed, just like formulating instructions for human evaluators.

\subsection{Story Generation}
Story generation is another NLG task with more emphasis on open-ended generation compared with text summarization, which also means for a given beginning of a story, various generated storylines and different plots could satisfy people. Therefore, story generation models are extremely challenging to evaluate. The automatic similarity-based metrics between the generated storylines and so-called references cannot fully evaluate the quality of the storylines since they do not consider creativity.

To show the effectiveness of ChatGPT as an NLG metric for the story generation task, we conduct experiments on OpenMEVA-ROC~\cite{guan-etal-2021-openmeva}. The OpenMEVA-ROC dataset manually annotates five model-generated storylines under the consideration of their overall quality.

\vspace{0.5ex}
\noindent \textbf{The Potentiality of ChatGPT.} As shown in Table~\ref{table:all_level_openmeva}, ChatGPT achieves the best performance in terms of all correlations, and significantly outperforms the second-best metric (\emph{i.e.}, PPL).
For example, ChatGPT (DA w/o ref) achieves 0.507 sample-level Spearman correlation, while PPL only achieves 0.324 sample-level Spearman correlation.
In addition, we also find that all similarity-based metrics (\emph{i.e.}, ROUGE-1, ROUGE-2, ROUGE-L, BERTScore and BARTScore) show their weak correlations with the human judgments.
This finding indicates that the ChatGPT evaluator has more powerful and reliable judgments on the open-ended and creative text generation tasks, where many diversified generated results could also be regarded as high-quality.

\vspace{0.5ex}
\noindent \textbf{The Impact of Different Prompt.} The results in Table~\ref{table:all_level_openmeva} also show the sensitivity of the correlation results led by the different prompts. For example, there are large performance gaps between ChatGPT (DA w/o ref) and ChatGPT (DA w/ ref). This finding is also consistent with that in text summarization (Section~\ref{subsec:4.3}).
More recently, some researchers also discuss the robustness of LLMs on different (adversarial) prompts~\cite{zhu2023promptbench}, and we think this under-explored LLM research direction deserves more research attention.

\subsection{Data-to-Text Generation}
Data-to-text generation aims at generating a fluent free-text description for a given structured table. We conduct experiments on BAGEL~\cite{mairesse-etal-2010-phrase} to show the effectiveness of the ChatGPT evaluator on data-to-text generation.

Table~\ref{table:dataset_level_bagel} shows the experimental results, where ChatGPT achieves competitive correlations compared with the previous state-of-the-art baselines, indicating its strong potentiality serving as a metric for data-to-text generation.
It is worth noting that we do not provide reference-free ChatGPT performance in terms of informativeness because informativeness in BAGEL is defined as ``whether the system generation contains all the information in the gold reference'', which also means that when evaluating informativeness the golden references must be given.

\section{Conclusion}
In this technical report, we explore a research question: ``\emph{Is ChatGPT a good NLG evaluator?}''. To this end, we design task-specific as well as aspect-specific prompts to guide ChatGPT to perform as an NLG metric. Experimental results on five widely-used meta-evaluation datasets, covering text summarization, story generation and data-to-text tasks, show the potentiality of ChatGPT as an NLG metric.
ChatGPT achieves the new state-of-the-art correlations (with human judgments) on SummEval and OpenMEVA meta-evaluation datasets, and obtains competitive results on NewsRoom and BAGEL datasets.

In addition, we also find that the lexical biases involved in the meta-evaluation datasets would influence the effectiveness of NLG metrics, and might lead to the limited performance of the ChatGPT evaluator. Besides, the performances of ChatGPT as an NLG evaluator are sensitive to the format of the prompt, for different tasks and aspects, the prompt should be carefully designed.

We believe that ChatGPT will exceed its current performance and provide a reliable NLG metric for the research community in the near future.

\section*{Limitations}
While we show that ChatGPT achieves state-of-the-art or competitive correlation with human judgments on various NLG tasks, there are limitations that provide avenues for future work: (1) ChatGPT's performance as an NLG metric is related to prompts, and future work could explore more powerful prompts to achieve better performance; (2) This preliminary report misses experiments on some mainstream NLG tasks, \emph{e.g.}, dialogue generation and report generation; (3) When we did the experiments, the OpenAI ChatGPT did not release the official API. Thus, we conducted the experiments on the ChatGPT website with default temperature, making the results difficult to reproduce. All experiments related to ChatGPT are conducted between February 24 to February 27, 2023; and March 17 to March 22.
(4) The experiments are only conducted on the English NLG meta-evaluation datasets, and future work could extend this method into other languages or cross-lingual scenes.
(5) The correlation between the ChatGPT evaluator and humans is also related to the quality and challenge of the corresponding meta-evaluation datasets. Our experiments are conducted on the traditional NLG meta-evaluation datasets (that appear before the LLM era). Recently, \citet{zeng2023evaluating} propose LLM-BAR, a challenging meta-evaluation benchmark to test the ability of an LLM evaluator. Future work could adapt our method to other challenging datasets and study the performance of the ChatGPT evaluator.

\section*{Acknowledgement}
We thank anonymous reviewers for their constructive suggestions and comments.

\bibliography{references}
\bibliographystyle{acl_natbib}

\appendix

\end{document}